# A Testbed for Cross-Dataset Analysis

Tatiana Tommasi, Tinne Tuytelaars, Barbara Caputo

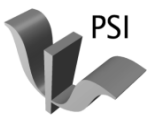

# A Testbed for Cross-Dataset Analysis

Tatiana Tommasi, Tinne Tuytelaars
KU Leuven
Belgium
firstname.lastname@esat.kuleuven.be

Barbara Caputo
Sapienza, University of Rome
Italy
caputo@ids.uniroma1.it

## Abstract

*Since its beginning visual recognition research has tried to capture the huge variability of the visual world in several image collections. The number of available datasets is still progressively growing together with the amount of samples per object category. However, this trend does not correspond directly to an increasing in the generalization capabilities of the developed recognition systems. Each collection tends to have its specific characteristics and to cover just some aspects of the visual world: these biases often narrow the effect of the methods defined and tested separately over each image set.*

*Our work makes a first step towards the analysis of the dataset bias problem on a large scale. We organize twelve existing databases in a unique corpus and we present the visual community with a useful feature repository for future research.*

## 1. Introduction

In the last two decades computer vision research has lead to the development of many efficient ways to describe and code the image content, and to the definition of several highly performing pattern recognition algorithms. In this evolution a key role was held by different image collections defined both as source of training samples and as evaluation instruments. The plethora of datasets obtained as legacy from the past, together with the modern increasing amount of freely available images from the Internet, has recently caught the attention of the computer vision community. Researchers started to elaborate over it mainly in two directions. On one side there is a growing interest for *large scale data* [7, 6, 30], moved by how to mine a huge amount of information and by how to use it to tackle difficult problems that were not solvable or not even thinkable before. On the other side there is the *dataset bias* problem [34, 18, 33]. Every finite image collection tends to be biased due to the acquisition process (used camera, lighting condition, etc.), preferences over certain types of background, post-processing elaboration (image filtering), or annotator tendencies (chosen labels). As a consequence the same object category in two datasets can appear visually different, while two different labels can be assigned to exactly the same image content. Moreover, not all the datasets cover the same set of classes, thus the definition of what an object "is not" changes depending on the considered collection.

Previous work has proposed initial solutions to evaluate the dataset bias or to overcome it, but the literature misses a standard testbed for large scale cross-dataset analysis. We believe that widening the attention from few shared classes to the whole dataset structures can reveal much about the nature of the biases, give insight on how to organize an unbiased image archive and allow to extend the experience gained by years of research on each image collection to the others. We organize several existing databases in a unique corpus and we present to the visual community a useful feature repository for future research.

## 2. Related Work

The existence of several data related issues in any area of automatic classification technology was first discussed by D. J. Hand in [15] and [16], while the first sign of peril in image collections was indicated in presenting the Caltech256 dataset [14] when the authors recognized the danger of learning ancillary cues of the dataset (characteristic image size) instead of intrinsic features of the object categories. However, only recently this topic has been officially put under the spotlight for computer vision tasks by A. Torralba and A. Efros [34]. Their work pointed out the idiosyncrasies of the existing image datasets: the evaluation of cross-dataset performance revealed that standard detection and classification methods fail because the uniformity of training and test data is not guaranteed.

This initial analysis of the *visual dataset bias* problem gave rise to a stream of works focusing on how to overcome the specific image collection differences and learn robust classifiers with good generalization properties. The proposed methods are mainly tested on binary tasks (object vs rest) where the attention is focused on categories like *car* or *person* which are common among six popular datasets: Sun09, LabelMe, PascalVOC, Caltech101, Imagenet, and MSRC [34]. A further group of three classes was also added to the original set (*bird*, *chair* and *dog*) defining a total of five object categories over the first four datasets listed before [18, 9]. A larger scale analysis in terms of categories was proposed in [33], considering also the problem of partial overlapping label sets among different datasets.

Together with the growing awareness about the characteristic signature of each existing image set, the related problem of *visual domain shift* has also emerged. In real life settings it is often impossible to control how the test images will differ from the training data: the two sets can be considered as belonging to two domains that need to be recomposed in a single one before applying any automatic pattern recognition method. Given a source and target image set with different marginal distributions, an efficient (and possibly unsupervised) solution is to learn a shared representation that eliminates the original distribution mismatch. Several methods based on subspace data embedding [13, 2], metric [29, 32] and vocabulary [28] learning have been presented. Some of these strategies have a long tradition in machine learning and natural language processing [3, 4, 5] and were reshaped for computer vision tasks. However, if in those areas of research the definition of a domain is well established, in computer vision referring only to the underlying marginal probability of the data may be not enough. First of all the nature of the mismatch among the distributions has not been fully analyzed, for instance it is not clear if the covariate shift assumption is realistic and some work have shown that the conditional probability distribution shift should be taken into consideration [24]. Moreover, when dealing with images, one of the main interests is in the motivation underling the visual aspect change. The causes of visual domain shift are numerous and although for some of them (lighting condition, image resolution [29]) may be relatively easy to define a robust image representation, for others (from pictures to paintings [31], time evolution [19]) recomposing the shift may be more difficult.

Despite their close relation, visual domain and dataset bias are not the same. Domain discovery approaches have shown that a single dataset may contain several domains [17] and on the other hand, a same domain may be shared across several datasets [12].

## 3. Collection Details

The standard first step of any computer vision research pipeline consists in collecting images for the task at hand. The divergent goals of different research lines can be accounted as one of the main reasons behind existing dataset bias. For our analysis we choose *object categorization* as the main general task and we consider 12 among the different inflections of this problem with the corresponding image collections created in the last years.

**ETH80** [22]. This dataset was created to facilitate the transition from object identification (recognize a specific given object instance) to image categorization (assign the correct class label to an object instance never seen before). It contains 8 categories and 10 toy objects for every category. Each object is captured against a blue background and it is represented by 41 images from viewpoints spaced equally over the upper viewing hemisphere.

**Caltech101** [11]. This dataset contains 101 object categories and was the first large scale collection proposed as a testbed for object recognition algorithms. Each category contain a different number of samples going from a minimum of 31 to a maximum of 800. The images have little or no clutter with the objects centered and presented in a stereotypical pose.

**Caltech256** [14]. Differently from the previous case the images in this dataset were not manually aligned thus the objects appear in several different poses. This collection contains 256 categories with a minimum of 80 and a maximum of 827 images.

**Bing** [1]. This dataset contains images downloaded from the Internet for the same set of 256 object categories of the previous collection. Text queries give as output several noisy images which are not removed, resulting in a weakly labeled collection. The number of samples per class goes from a minimum of 197 to a maximum of 593.

**Animals with Attributes (AwA)** [21]. This is a database of 50 animals categories, containing a total of 30475 images. Each class is associated to a 85-element vector of numeric attribute values that indicate general characteristics shared between different classes. The animals appear in different pose and at different scales in the images.

**a-Yahoo** [10]. As the previous one, this dataset was collected to explore attribute descriptions. It contains 12 object categories with a minimum of 48 and a maximum of 366 samples per class.

**MSRCORID** [26]. The Microsoft Research Cambridge Object Recognition Image Database contains a set of digital photographs grouped into 22 categories spanning over objects (19 classes) and scenes (3 classes).

**PascalVOC2007** [8]. The Pascal Visual Object Classes dataset contain 20 object categories and a total of 14319 images. Each image depicts objects in realistic scenes and may contain instances of more than one category. This dataset was used as testbed for object recognition and detection challenges over several years.

**SUN** [36]. This dataset contains a total of 142165 pic-tures[1] and it was created as a comprehensive collection of annotated images covering a large variety of environmental scenes, places and objects. Here the objects appears at different scales and positions in the images and many of the instances are partially occluded making object recognition and categorization extremely challenging.

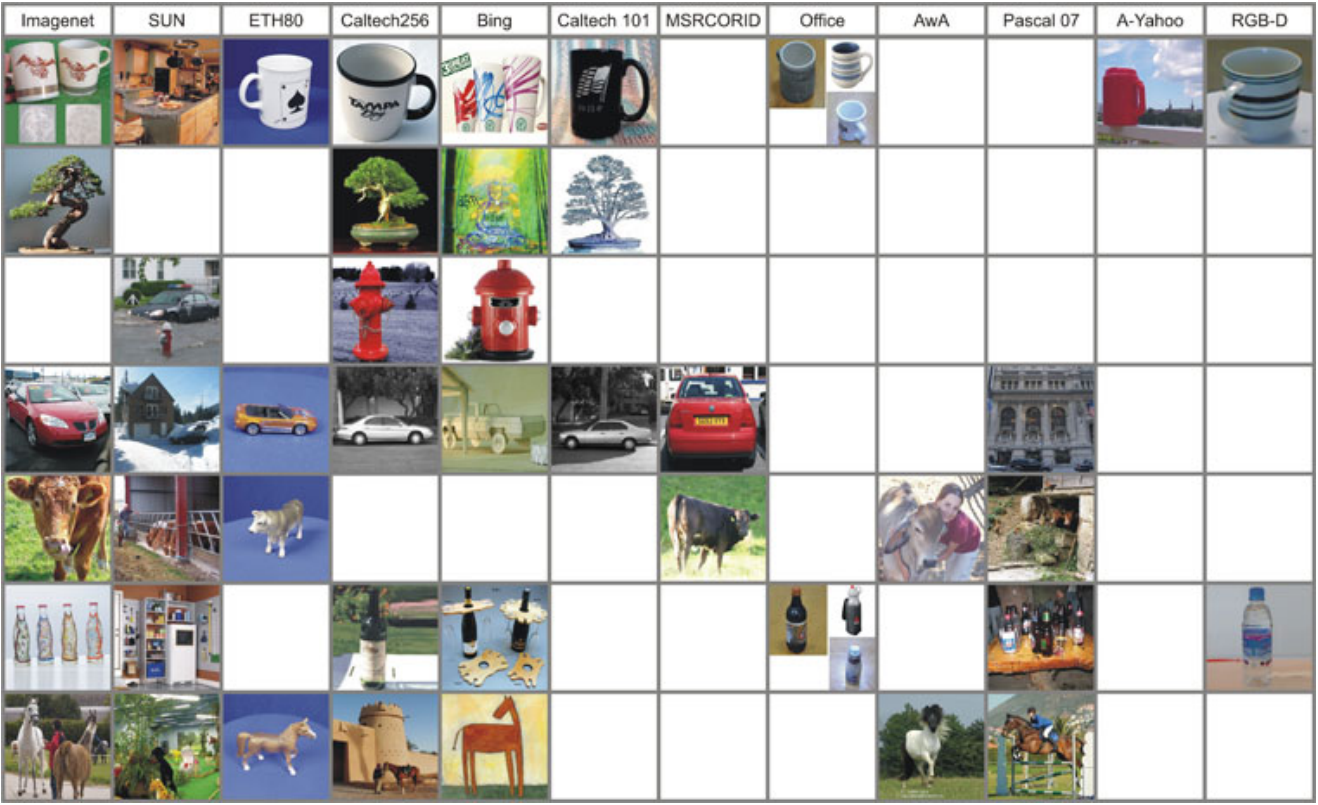

Figure 1. We show here one example extracted from each of the 12 datasets for 7 object categories, one per row: mug, bonsai, fire hydrant, car, cow, bottle, horse. Here we consider cup and mug as the same class as well as bottle, wine-bottle and water-bottle. The empty positions indicate that the dataset associated to that column does not contain the corresponding class. For Office we show one sample per domain.

**Office** [29]. The Office dataset contains images of 31 object classes over three domains: the images are either obtained from the Amazon website, or acquired with a high resolution digital camera (DSLR), or taken with a low resolution webcam. The collection contains a total of 4110 images with a minimum of 7 and a maximum of 100 samples per domain and category.

**RGB-D** [20]. This dataset is similar in spirit to ETH80 but it was collected with a Kinect camera, thus each RGB image is associated to a depth map. It contains images of 300 objects acquired under multiple views and organized into 51 categories.

---

[1]Here we consider the latest version available at http://labelme.csail.mit.edu/Release3.0/Images/users/antonio/static_sun_database/ and the list of objects reported at http://groups.csail.mit.edu/vision/SUN/.

**Imagenet** [7]. At the moment this collection contains around 21000 object classes organized according to the Wordnet hierarchy.

Overall, the images have different resolution and dimensions. Some of them are in jpg format, while others are png files. Figure 1 depicts some samples extracted from different datasets, giving an idea of the variability across them. There are two main challenges that must be faced when thinking to organize and use all these data collections at once.

**Merging**: finding a way to properly compose the datasets in a single corpus turns out to be quite difficult. Even if each image is labeled with an object category name, the class alignment is tricky due to the use of different words to indicate the very same object, for instance *bike* vs *bicycle* and *mobilephone* vs *cellphone*. Sometimes the different nuance of meaning of each word are not respected: *cup* and *mug* should indicate two different objects, but the images are often mixed; *people* is the plural of *person*, but images of this last class (PascalVOC07) often contain more than one subject. Moreover, the choice of different ontology hierarchical levels (*dog* vs *dalmatian* vs *greyhound*, *bottle* vs *water-bottle* vs *wine-bottle*) complicates the combination.

Figure 2. Stack histogram showing the number of samples for each of the Caltech256 classes over four datasets.

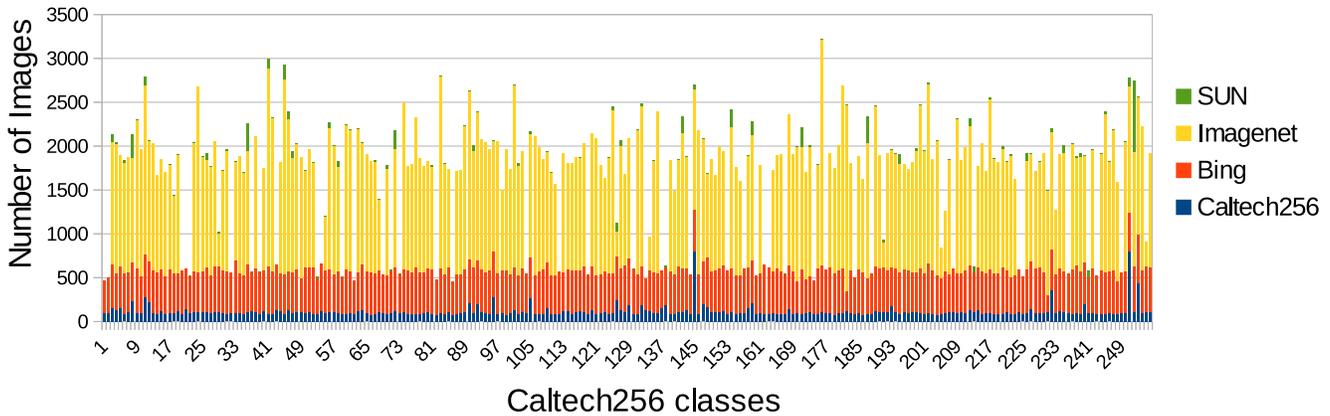

Figure 3. (Best viewed in color and magnification) Three cases of Imagenet categories. Left: some images in class *chess* are wrongly labeled. Middle: the class *planchet* or coin blank contains images that can be more easily labeled as *coin*. Right: some images in the class *truncated pyramid* do not contain a pyramid.

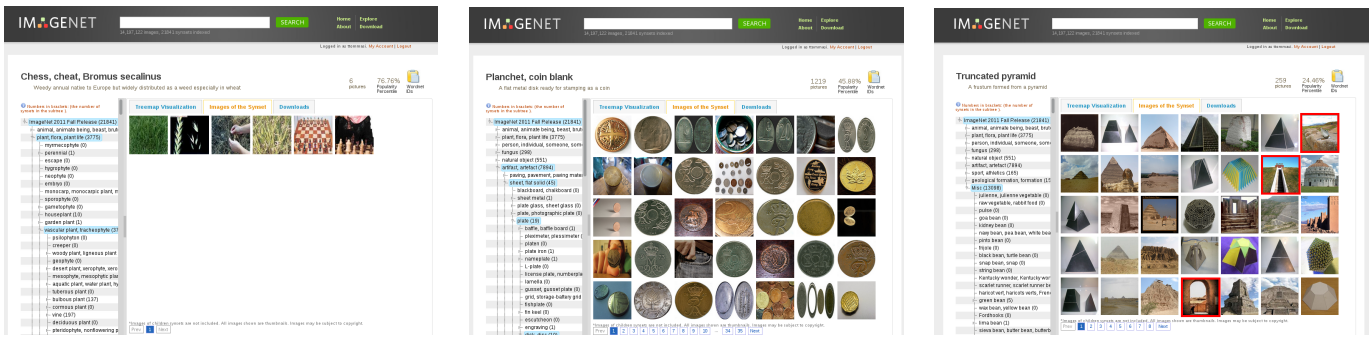

Figure 4. Three categories with labelling issues. The class *bat* has different meanings both across datasets and within a dataset. A *saddle* can be a seat to ride a horse or a part of a bicycle. A *skateboard* and a *snowboard* may be visually similar, but they are not the same object

**Representation**: although many studies have been performed on each dataset separately and several representations have been tested and publicly released, there is no previous work that extracted and calculated exactly the same feature for all the images of several dataset at a large scale[2].

We propose to tackle these two issues by defining two data setups and by extracting several feature representations. All the details are given in the following subsections.

### 3.1. Data Setup

It is possible to define two different combinations of data.

**Dense set**. Among the considered datasets, the ones with the highest number of categories are Caltech256, Bing, SUN and Imagenet. Infact the last two are open collections progressively growing in time. Overall they share 114 categories: some of the 256 object categories are missing at the moment in Imagenet but they are present in SUN (*e.g.* 053.desk-globe, 071.fire-hydrant) and vice-versa (*e.g.* 024.butterfly, 165.pram). The histogram bars in Figure 2 show the statistics of the overlapping. Out of this shared group, 40 classes (see Figure 5) contain more than 20 images per dataset and we selected them as our dense cross-dataset setup. We remark that each image in SUN is annotated with the list of objects visible in the depicted scene: we consider an image as a sample of a category if the category name is in the mentioned list.

---

[2] Some existing repositories: for Caltech 101 and 256 http://files.is.tue.mpg.de/pgehler/projects/iccv09/, five classes over four datasets http://undoingbias.csail.mit.edu/, classemes features for Bing and Caltech 256 http://vlg.cs.dartmouth.edu/projects/domainadapt/

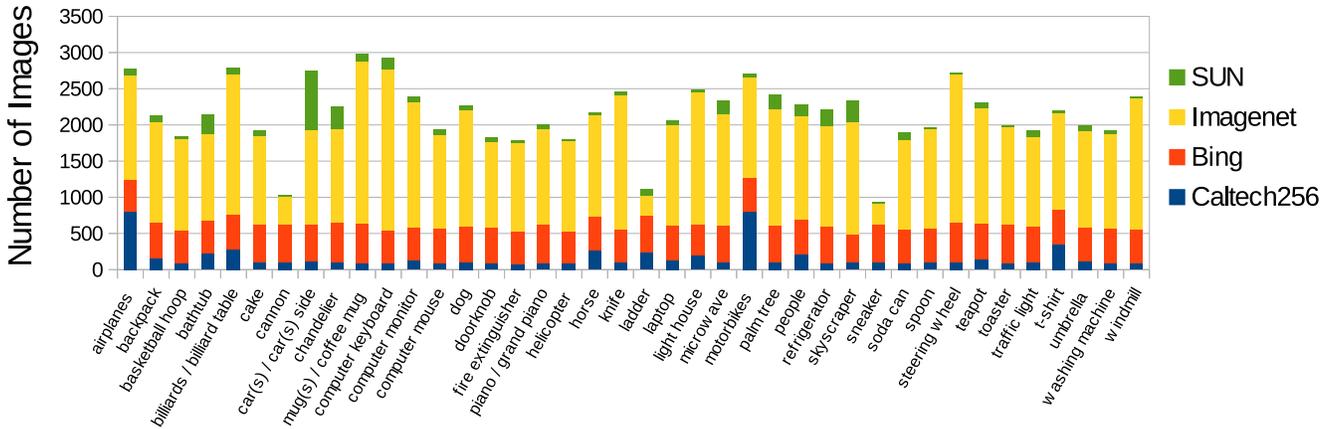

Figure 5. Stack histogram showing showing the number of images per class of our cross-dataset dense setup.

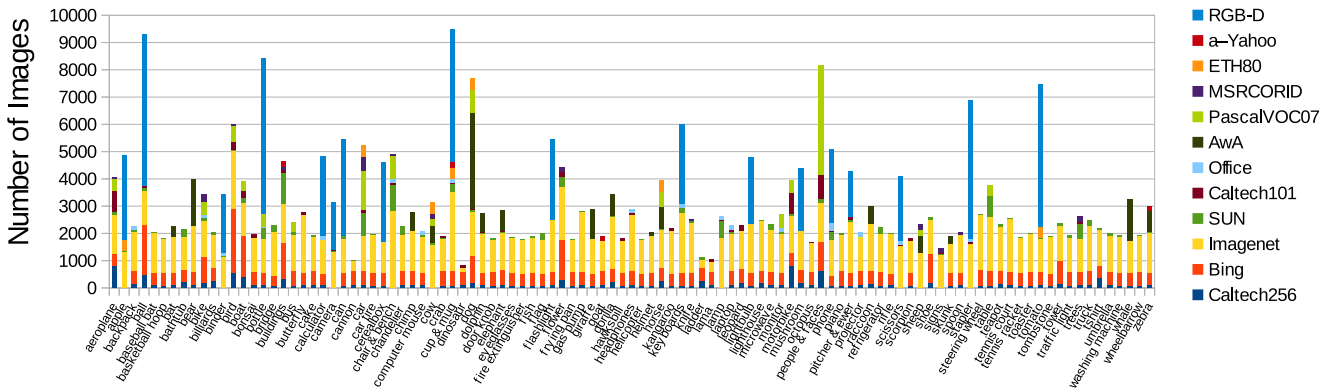

Figure 6. Stack histogram showing the number of images per class of our cross-dataset sparse setup.

In the alignment process we came across few peculiar cases. Figure 3 shows the samples of three classes in Imagenet. The category *chess board* does not exist at the moment, but there are three classes related to the word *chess*: chess master, chessman or chess piece, chess or cheat or bromus secalinus (we use "or" here to indicate different labels associated to the same synset). This last category contains only few images but some of them are not correctly annotated. The categories *coin* and *pyramid* are still not present in Imagenet. For the first, the most closely related clas is *planchet* or *coin blank*, which still contain many example of what would be commonly named as a coin. For the second, the most similar *truncated pyramid* contains images of some non-truncated pyramids as well as images not containing pyramids at all. In general, it is important to keep in mind that several of the Imagenet pictures are weakly labeled, thus they cannot be considered as much more reliable than the corresponding Bing images. Imagenet users are asked to clean and refine the data collection by indicating whether an image is a typical or wrong example.

Over all the four datasets we noticed that the word *bat* usually indicates the flying mammal except than in SUN where it refers to the baseball and badminton bat. A *saddle* in Caltech256 is the supportive structure for a horse rider, while in SUN it is a bicycle seat. Tennis shoes and sneakers are two synonyms associated to the same synset in Imagenet, while they correspond to two different classes in Caltech256. In SUN, there are two objects annotated as skateboards, but they are infact two snowboards. Some examples are shown in Figure 4.

**Sparse set**. This setup is obtained by searching over all the datasets for the categories with a minimum of 20 samples and shared by at least four collections. We allow a lower number of samples only for the classes shared by more than four datasets (*i.e.* from the fifth dataset on the images per category may be less than 20). These conditions are satisfied by 105 object categories in Imagenet overlapping with 95 categories of Caltech256 and Bing, 89 categories of SUN, 34 categories of Caltech101, 17 categories of Office, 18 categories of RGB-D, 16 categories of AwA and PascalVOC07, 13 categories of MSRCORID, 7 categories of ETH80 and 4 categories of a-Yahoo. To obtain this combination we choose some high level concept corresponding to a node in the Wordnet hierarchy with several leaves. For instance, *bird* covers humming bird, pigeon, ibis, flamingo, flamingo head, rooster, cormorant, ostrich and owl; *boat* covers kayak, ketch, schooner, speed boat, canoe and ferry. The histogram in Figure 6 shows the defined sparse set and the number of images per class: the category *cup & mug* is shared across nine datasets, resulting the most popular one.

### 3.2. Representation

The choice of image representation has an important role in smoothing the domain shift and the dataset bias. We consider here several features.

**Dense SIFT**. This is one of the most widely adopted image representation. We followed the same protocol proposed in the development kit[3] of the Large-Scale Visual Recognition Challenge of 2010 (ILSVRC 2010). Each image is resized to have a max size length of no more than 300 pixels and SIFT descriptors are computed on 20x20 overlapping patches with a spacing of 10 pixels. Images are further downsized (to 1/2 and 1/4 of the side length) and more descriptors are computed. We provide both the raw descriptors and the Bag of Words representation obtained by clustering a random subset of 10 millions SIFT vectors to form a visual vocabulary of 1000 visual words.

**Object-Classemes**. Several recent papers have shown that the classification output of a bunch of object or concept models can be used as an effective descriptor [23, 25]. The original classeme representation presented in [35] was based on a set of weakly labeled visual concept associated to images downloaded from a search engine. Here we follow the same idea but we start from a set of 1000 object classes (ILSVRC 2010) to create linear SVM models over Fisher Vectors. Any image is then represented by the obtained 1000 set of classification output values.

### 4. Conclusion

In this report we discussed the challenges faced when aligning twelve existing image datasets and we propose two data setups that can be used as testbed for cross-dataset analysis. We extracted several descriptors from the images and we created a public feature repository that can be useful for future research.

We consider this as the first step of a wider project (official webpage: https://sites.google.com/site/crossdataset/) that will continue by running several cross-dataset classification tests. We believe that exploring the common information and the specific aspects of each collection will not only reveal more about the dataset bias problem, but it will also allow to transfer knowledge across datasets. Some of the images are associated with attributes, others with point clouds and depth information: both may be shared with other samples which miss those cues. Combining scene datasets with object-centric ones may be used to associate the typical environment where an object can be found with a more detailed knowledge of its appearance and visual variability. Moreover, we plan to study the relation between our manual alignment process and what could be automatically achieved by leveraging over the entry-level categories proposed in [27].

---

[3] www.image-net.org/download-features